\documentclass[conference]{IEEEtran}
\IEEEoverridecommandlockouts
\usepackage{cite}
\usepackage{amsmath,amssymb,amsfonts}
\usepackage{algorithmic}
\usepackage{graphicx}
\usepackage{textcomp}
\usepackage{xcolor}
\usepackage{multirow}
\usepackage{balance}
\usepackage{orcidlink}

\def\BibTeX{{\rm B\kern-.05em{\sc i\kern-.025em b}\kern-.08em
    T\kern-.1667em\lower.7ex\hbox{E}\kern-.125emX}}

\begin{document}

\title{Sentiment Analysis of Twitter Posts on Global Conflicts \\
}

\author{\IEEEauthorblockN{Ujwal	Sasikumar}
	\IEEEauthorblockA{\textit{School of Workforce Development} \\
		\textit{Conestoga College ITAL}\\
		Kitchener ON, Canada \\
		ujusasikumar@gmail.com}
~\\
\and
\IEEEauthorblockN{ANK Zaman$\textsuperscript{\orcidlink{0000-0001-7831-0955}}$}
\IEEEauthorblockA{\textit{Dept. of Physics and Computer Science} \\
\textit{Wilfrid Laurier University}\\
Waterloo ON, Canada \\
azaman@wlu.ca}
~\\
\and
\IEEEauthorblockN{Abdul-Rahman	Mawlood-Yunis}
\IEEEauthorblockA{\textit{Dept. of Physics and Computer Science} \\
	\textit{Wilfrid Laurier University}\\
	Waterloo ON, Canada \\
	amawloodyunis@wlu.ca}
~\\
\and
\IEEEauthorblockN{Prosenjit Chatterjee$\textsuperscript{\orcidlink{0000-0003-1169-4717}}$}
\IEEEauthorblockA{\textit{Dept. of Computer Science and Cyber Security} \\
	\textit{Southern Utah University}\\
	Cedar City, UT, USA \\
	prosenjitchatterjee@suu.edu}
	*Corresponding author
~\\
}

\maketitle

\begin{abstract}
Sentiment analysis of social media data is an emerging field with vast applications in various domains. In this study, we developed a sentiment analysis model to analyze social media sentiment, especially tweets, during global conflicting scenarios. To establish our research experiment, we identified a recent global dispute incident on Twitter and collected around 31,000 filtered Tweets for several months to analyze human sentiment worldwide. In this research, the training models comprised of Naïve Bayes Algorithm (NBA) and Neural Networks (NN). The trained model achieved 80\% and 100\% accuracy in predicting the sentiment of the tweets using Naive Bayes and Neural Network models, respectively. A confusion matrix was created to evaluate the model’s performance, indicating that the model correctly classified many positive and negative tweets. This study contributes to developing sentiment analysis models for analyzing social media data related to various events and topics. The findings can be helpful for researchers and practitioners interested in understanding public sentiment about the country-wide conflict on social media. In this document, Twitter is referring to its new name X.
\end{abstract}


\section{Introduction}\label{int}
The complex geopolitical war between countries, which has enormous worldwide repercussions, has drawn much interest on social media platforms, particularly Twitter. This study thoroughly investigates the opinions posted on Twitter about the current warfare in the area. This work examines a large corpus of tweets using cutting-edge NLP and machine-learning (ML) approaches to find the prevailing sentiment trends, viewpoints, and feelings around this complex conflict. Modern sentiment analysis methods and procedures are used in our study to classify Twitter corpus into positive, negative, and neutral sentiment categories. From this analysis, we want to learn how people, groups, and the global community view and respond to the international wartime. This work also investigates temporal trends and regional differences in sentiment expression, offering information on how public opinion transformed throughout the conflict. 

%

A valuable resource for policymakers, researchers, and the general public looking to understand the tones~\cite{ZAVATTARO2015333} of public opinion and sentiment surrounding this ongoing global issue is provided by this study, which contributes to a deeper understanding of the social and emotional aspects of the international War between two countries. In the end, the findings demonstrate the influence social media has on influencing and reflecting public opinion during times of global crisis.
Social media has become an increasingly important source of information during major events and crises, including wars and conflicts~\cite{kumar2021social}. Twitter has been extensively used among social media platforms to share news, opinions, and emotions about ongoing events~\cite{kwon2018prominent}. The recent global conflict in the year 2022-23 between multiple countries is one such event that has garnered significant attention on social media, with users from around the world expressing their views and sentiments~\cite{liu2015sentiment, lu2020sentiment}. This study aims to build a sentiment analysis model to determine the sentiment of tweets during the said war posted by independent country (not involved in any war) users.
Sentiment analysis is a subfield of natural language processing (NLP) that involves using computational techniques to extract and quantify subjective information from text~\cite{liu2015sentiment}. NLP is a rapidly growing field of artificial intelligence that focuses on enabling computers to understand, interpret, and generate human language~\cite{jurafsky2019speech}. By applying NLP techniques, we aim to develop a model that can accurately classify tweets as positive, negative, or neutral based on the sentiment expressed in the text.
To achieve this goal, the NBA and NNs are popular ML techniques extensively used for text classification~\cite{rish2001empirical}. The algorithms learn the probability of each object’s features and classify them into several predetermined categories. Supervised learning is the approach to train our model, using a set of annotated tweets to identify the relationship between the tweet’s features and its corresponding sentiment. The main objective of this study is to evaluate the accuracy of the developed model in classifying the sentiment of tweets related to the international war posted by independent country’s users. The results of this study could have significant implications for understanding the impact of social media on public opinion and sentiment during times of global conflict.
Researchers have used a variety of approaches and technology in recent years to do sentiment analysis on Twitter data related to the war-time conflict scenario. This literature review delves into the most important studies and conclusions.
Serpil~\cite{aslan2023deep} examines sentiment analysis in the context of the Ukraine situation. It analyses public 
opinion using Twitter data and exploring users’ emotional responses to any war, illuminating how emotions change over time.
The rest of the paper is organized as follows: Section II describes the dataset used in this research, Section III represents the technical details of the implementation, Section IV represents results, and Section V describes the concluding remarks of this paper.

\section{Dataset}\label{d}
The following sections present the data collection and preprocessing steps to prepare the dataset for evaluation.

\subsection{Data Collection}
The data used for building the sentiment analysis model was collected from Canadian independent country’s users' tweets during the international wartime in the year 2022. The tweets were obtained using the Twitter API~\cite{snscrape_github} and stored in a CSV file. Additionally, we utilized Snscrape, a Python-based web scraper, to gather information from Twitter for instances that did not require API use.
We leveraged the Naïve Bayes Classifier, a popular machine learning algorithm, to analyze the collected tweets. Specifically, we utilized the Scikit-learn (sklearn) library~\cite{pedregosa2011scikit}~\cite{scikit-learn_website}, which is a free open-source machine-learning library for the Python programming language. Scikit-learn offers a variety of features such as classification, regression, and clustering, and we employed it for classification purposes.
Each tweet in our dataset was labeled as positive, neutral or negative based on the sentiment expressed in the text. To accomplish this, we used the TextBlob library~\cite{textblob_docs}, which assigns a sentiment score ranging from (-1) to (+1) to each tweet, with (-1) representing the most negative sentiment and (+1) representing the most positive. In addition, we filtered out tweets that did not contain relevant information to ensure the quality and validity of our dataset.

\subsection{Data Preprocessing}
Data preprocessing is critical to any ML and/or NLP activity. It involves transforming raw data into a format that ML algorithms can easily analyze and understand. Our study used data preprocessing on the twitter dataset to prepare the data for sentiment analysis.
The first step in data preprocessing was removing irrelevant data, such as numbers and punctuation marks.
To achieve the preprocessing goal, regular expression was used in identifying unwanted text for elimination~\cite{alvarez2017data}. In NLP research, frequently used words are called stop words~\cite{6093315} and are removed as they do not have significance in the context of sentiment analysis. This step was performed using the Natural Language Toolkit (NLTK) library, which provides access to predefined stop words for the English text~\cite{bird2009natural}.
In our second step, the text was converted to lowercase to ensure consistency in the dataset and avoid any inconsistencies arising from uppercase and lowercase characters in the text data~\cite{hutto2014vader}. Stemming~\cite{porter1980algorithm} was also utilized to reduce each word to its base or root form to further reduce the dimensionality of the dataset, making it easier to analyze and understand.
Parts-of-speech (POS) tagging and chunking were also employed to identify the grammatical structure of each sentence. This technique involved assigning POS tags, such as nouns, verbs, adjectives, or adverbs, to each word in the text data. The words were then processed together into adjacent words in a sentence into phrases based on their POS tags. The data processing phase was carried out using several libraries, including Pandas for reading the CSV file, NumPy for numerical computations, Seaborn and Matplotlib for data visualization, and Scikit-learn for ML algorithms~\cite{pedregosa2011scikit}. A popularly available cloud server platform was utilized to conduct the experiment.
The preprocessing techniques used in this study helped clean and transform the tweets dataset into a format easily analyzed and understood by ML algorithms. This step was critical in ensuring the accuracy and effectiveness of the sentiment analysis.

\subsection{Data Visualization}
\begin{figure}[!ht]
	\includegraphics[scale=.43]{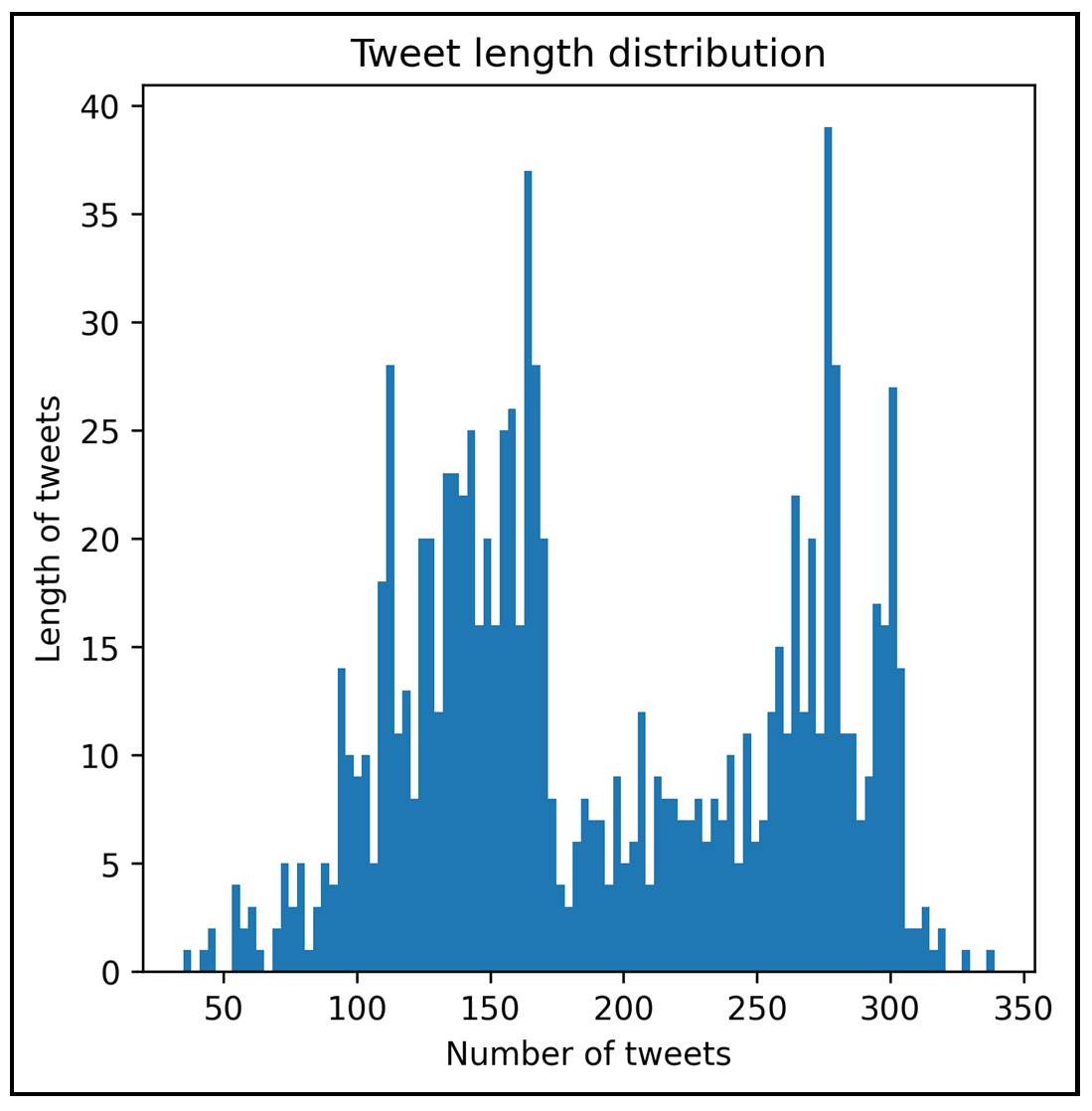}
	\centering
	\caption{Tweet length distribution}
	\label{fig:tl}
\end{figure}

After performing the data cleaning, we visualized the cleaned data for further processing. Figure~\ref{fig:tl} displays the sample length of tweets of the dataset.

\begin{figure}[!ht]
	\includegraphics[scale=.33]{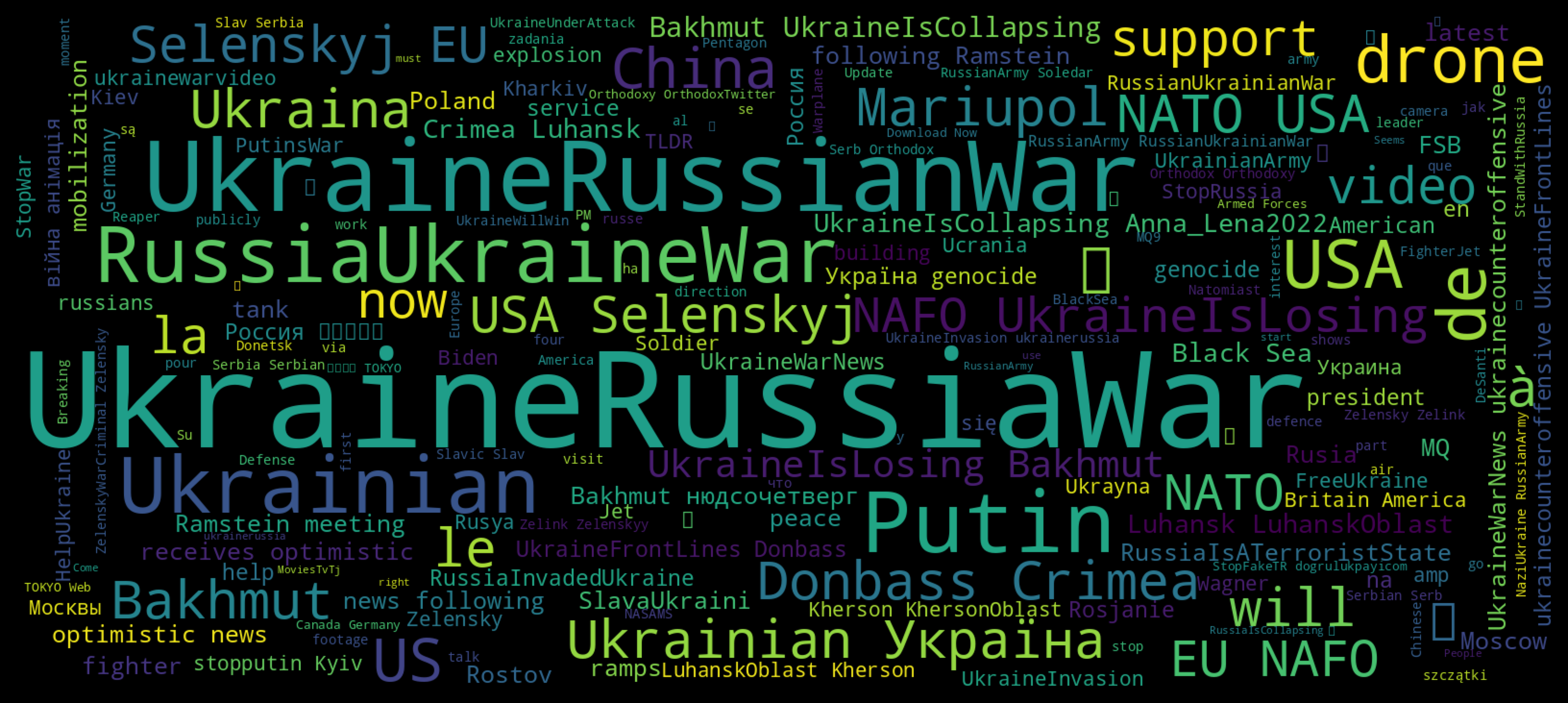}
	\centering
	\caption{Word cloud for neutral tokens}
	\label{fig:nut}
\end{figure}

Figure~\ref{fig:nut} represents the neutral tweet words from the dataset using a word cloud.

Figure~\ref{fig:pcloud} represents the positive tweet words from the dataset using a word cloud.

\begin{figure}[!ht]
	\includegraphics[scale=.33]{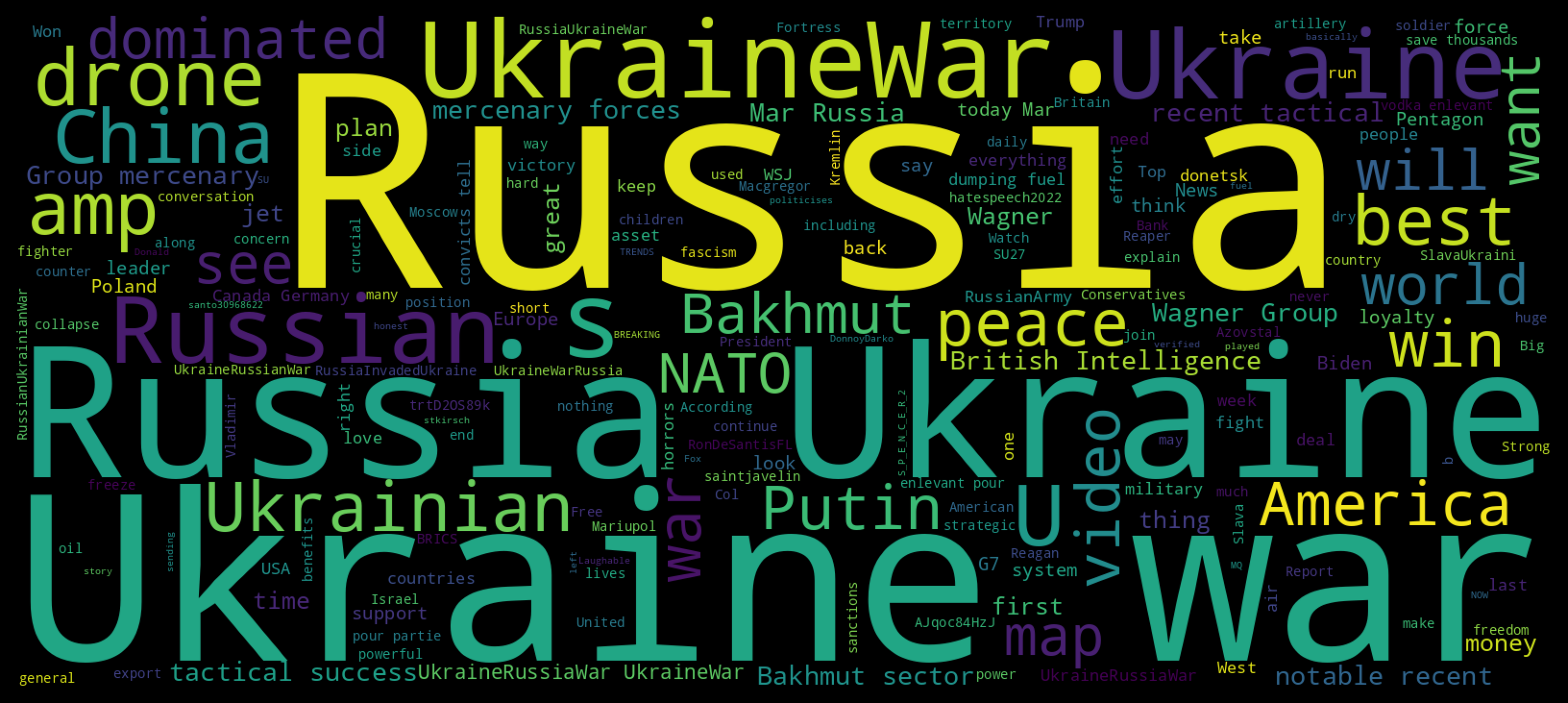}
	\centering
	\caption{Word cloud for positive tokens}
	\label{fig:pcloud}
\end{figure}

Figure~\ref{fig:pcloud} represents the negative tweet words from the dataset using a word cloud.

\begin{figure}[!ht]
	\includegraphics[scale=.33]{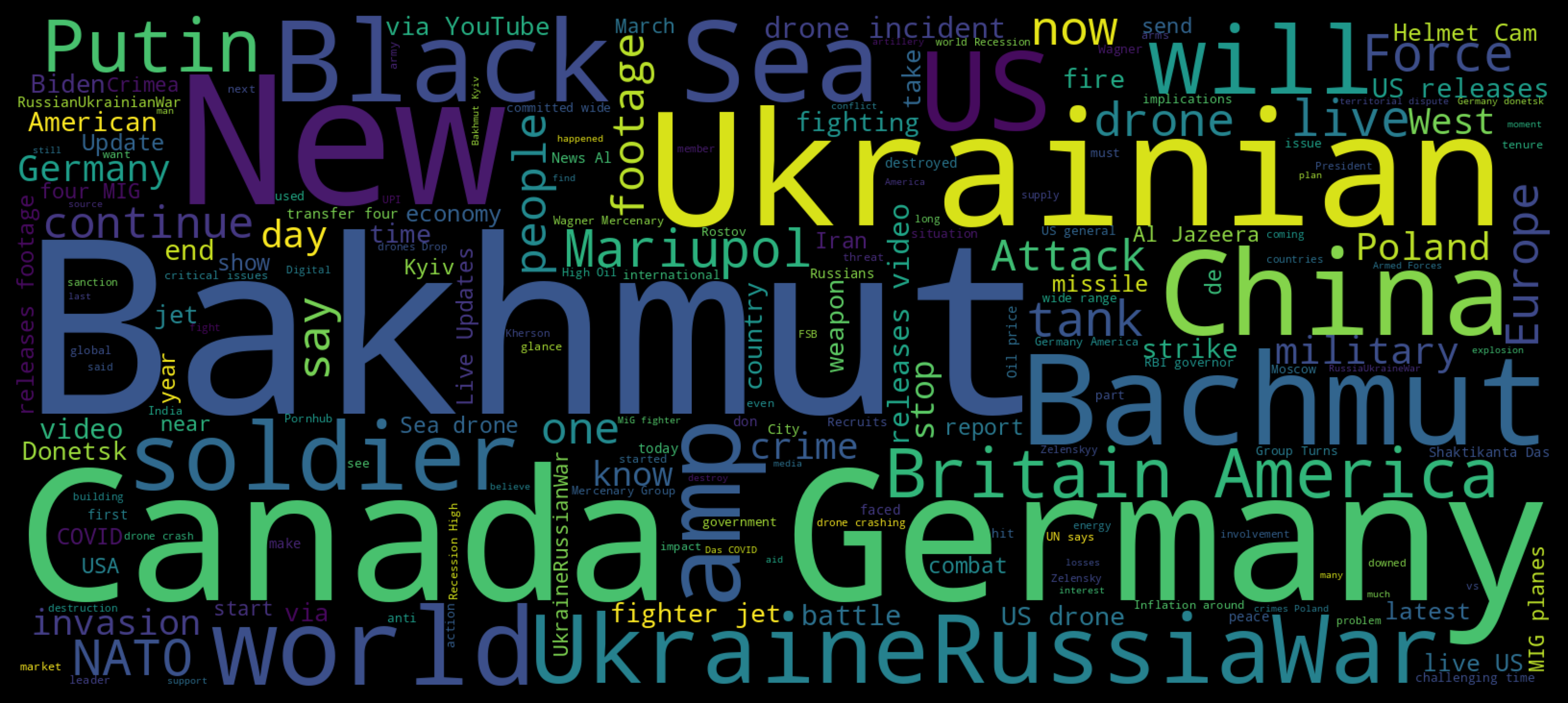}
	\centering
	\caption{Word cloud for negative tokens}
	\label{fig:pcloud}
\end{figure}

\section{Implementation}\label{imp}
The subsequent sections illustrate the process of building and training models to evaluate a sentiment dataset.

\subsection{Naïve Bayes Classifier Model}\label{nb}
We used the Naive Bayes (NB)~\cite{nb} classifier, a popular machine learning algorithm as a text classifier, to train our sentiment analysis model. The Naive Bayes classifier selects the class with the highest probability by first calculating the probability of each class based on the input features [16]. We split the dataset into training and testing sets, utilizing 80\% for training and 20\% for testing purposes.

\begin{figure}[!ht]
	\includegraphics[scale=.36]{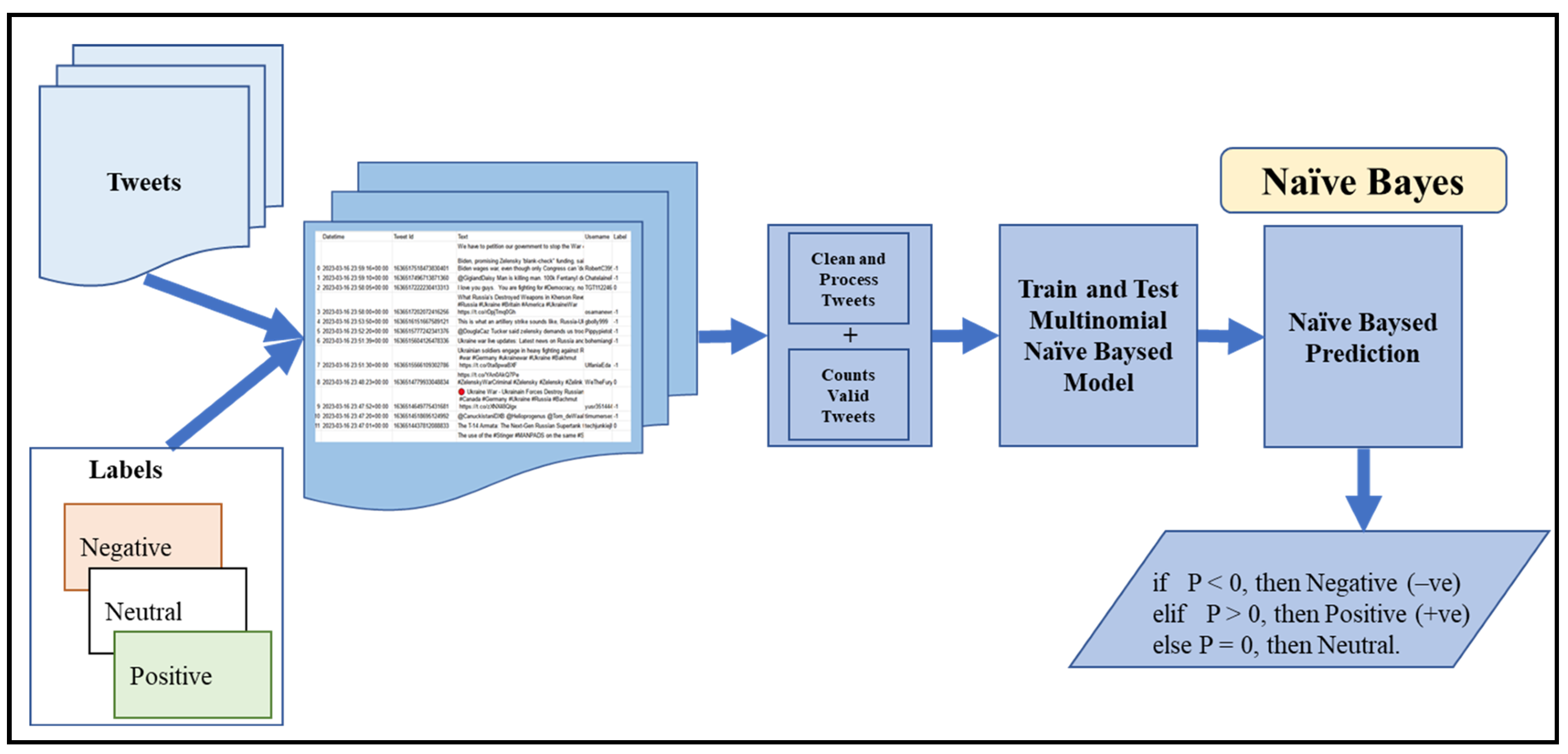}
	\centering
	\caption{Naïve Bayes Classifier Model}
	\label{fig:nbd}
\end{figure}

Figure~\ref{fig:nbd} shows the NB classifier model, and its processing steps.

For the purpose of ML data preprocessing we removed punctuations and stopwords from the tweet text.
We used the CountVectorizer~\cite{text2token} to convert the text data into a bag-of-words format, which represents the occurrence of each word in each tweet. The resulting matrix was split into features and labels. The Naive Bayes classifier was then trained using the training data, and its accuracy was evaluated using the testing data. To evaluate the performance of the model, we generated a confusion matrix and classification report. The confusion matrix shows the number of True Positives (TP), True Negatives (TN), False Positives (FP), and False Negatives (FN), while the classification report shows metrics such as precision, recall, and f1 score for each class. 

Precision is the ratio of true positives (TP) to the sum of true positives and false positives (FP). Recall: This is the ratio of true positives (TP) to the sum of true positives and false negatives (FN). f1-Score is the harmonic mean of precision and recall and is used as a single metric that balances the trade-off between precision and recall. The F1score ranges from 0 to 1, with 1 being the best possible score.

The formal definitions of the mentioned terms are given below:
\vspace*{2mm}
\par
$Accuracy = \frac{TP+TN}{TP+TN+FP+FN}$
\newline
\par

Here in this case, Precision is the ratio of true positives (TP) to the sum of true positives and false positives (FP).
\newline
\par
$Precision = \frac{TP}{TP+FP}$
\newline
\par
Recall: This is the ratio of true positives (TP) to the sum of true positives and false negatives (FN). 
\newline
\par
$Recall = \frac{TP}{TP+FN}$
\newline
\par

F1\-score is the harmonic mean of precision and recall and is used as a single metric that balances the trade-off between precision and recall. The f1\-score ranges from 0 to 1, with 1 being the best possible score. 
\newline
\par
$F1 = \frac{2*Precision*Recall}{Precision+Recall} = \frac{2*TP}{2*TP+FP+FN}$
\newline
\par

\subsection{Neural Networks}\label{nn}
We also experimented with a deep Neural Networks (NN) learning model for sentiment analysis using the Keras library. We defined a preprocessed text function to remove HTML tags, special characters, and stopwords from the tweet text. The Tokenizer was used to convert the text data into sequences of integers. The training and testing datasets were created using the train test split function, and the model was trained using the training data and evaluated using the testing data. The tokenizer was saved as JSON format for future use.

\begin{figure}[!ht]
	\includegraphics[scale=.35]{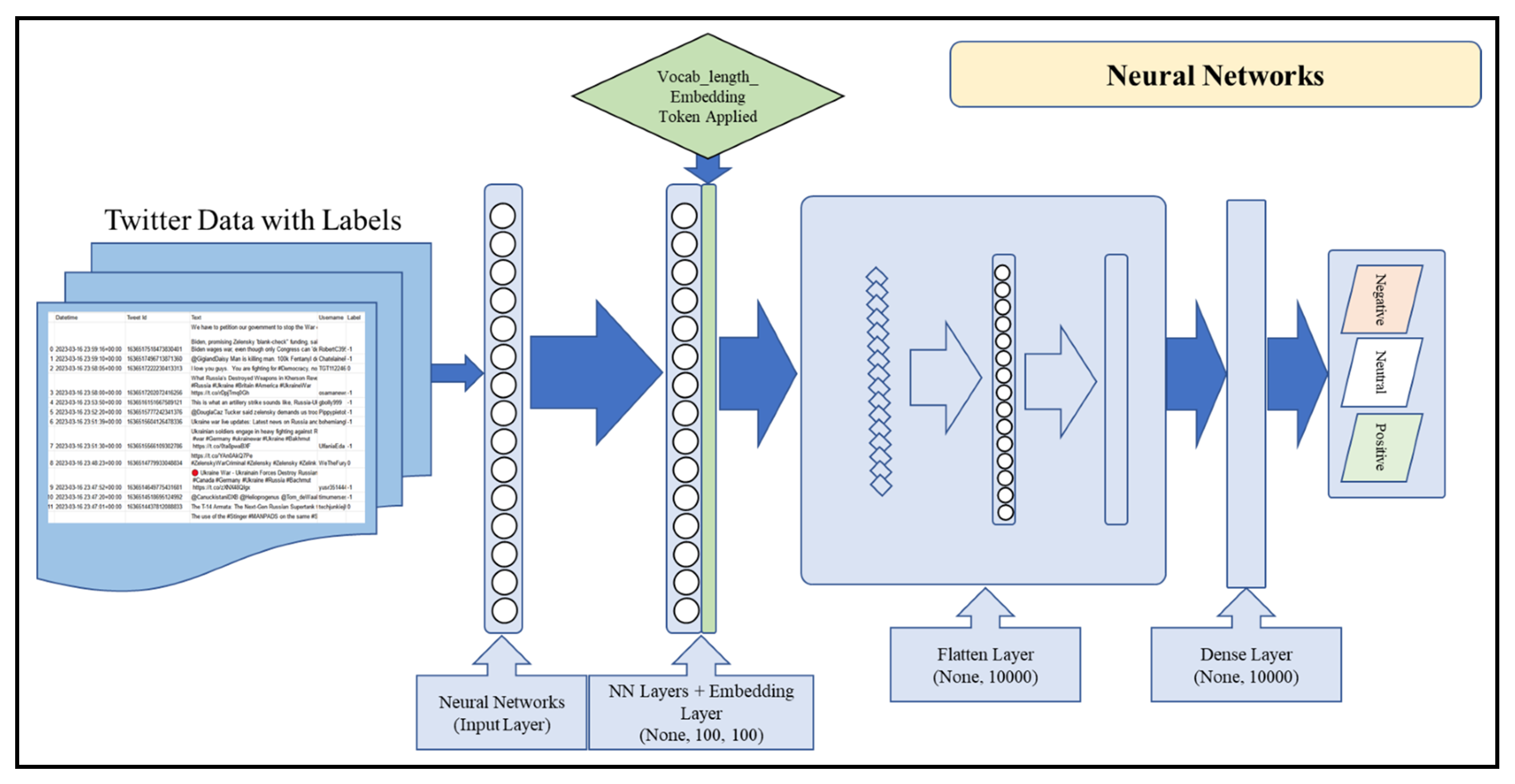}
	\centering
	\caption{Neural Network Model}
	\label{fig:nnd}
\end{figure}

Figure~\ref{fig:nnd} shows the NN model, and its processing steps.

\begin{figure}[!ht]
	\includegraphics[scale=.34]{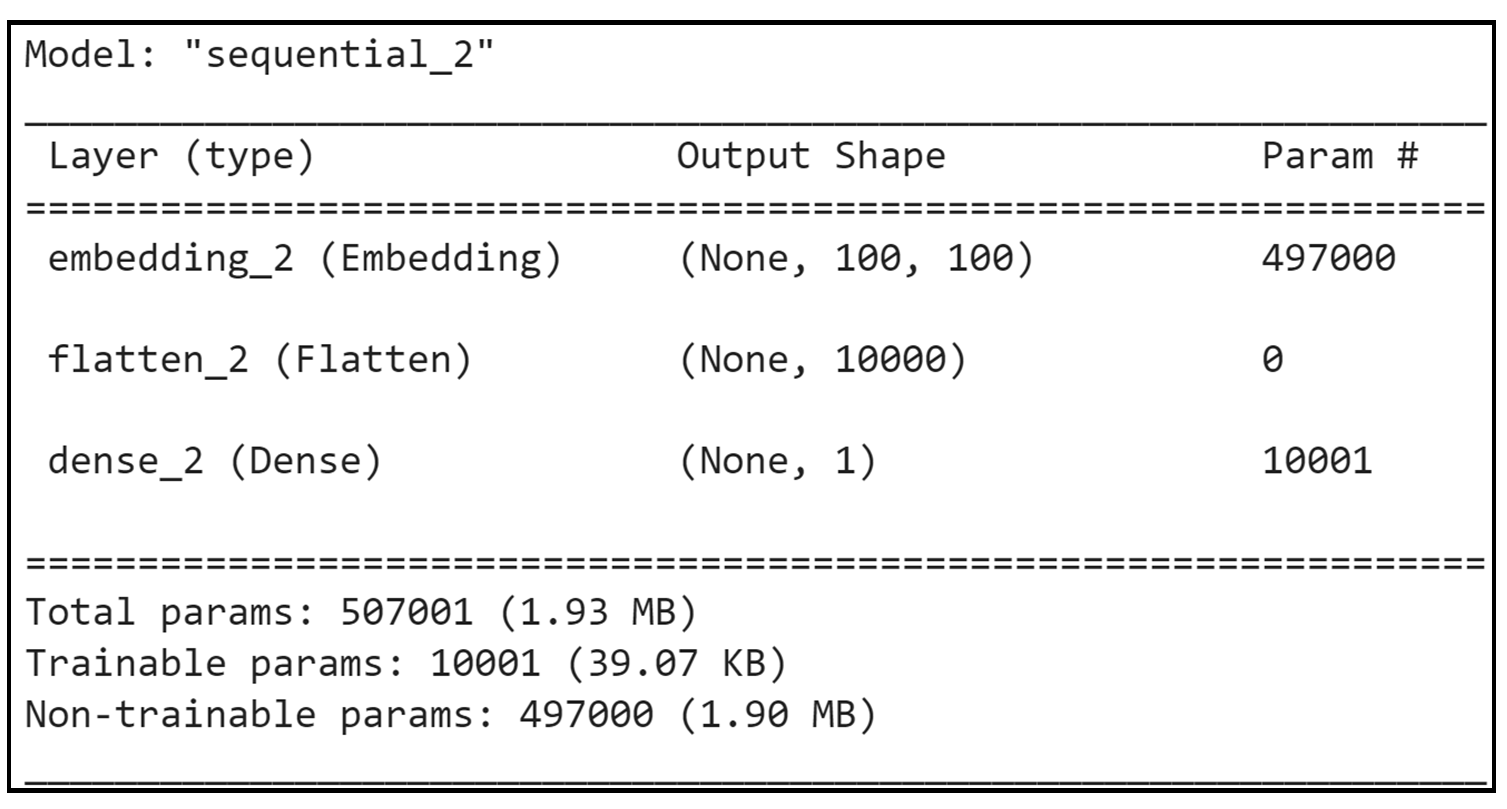}
	\centering
	\caption{Summary of the NN model }
	\label{fig:nn}
\end{figure}

Training a sentiment analysis model involves several essential steps, including data preprocessing, splitting the data into training and testing sets, and evaluating the model’s performance using various metrics. By following these steps, a robust and accurate sentiment analysis model using NN has developed to analyze large volumes of text data for various applications. Figure~\ref{fig:nn} presents the summary of the NN model’s layers.

\section{Result and Discussion}\label{resd}

This study aimed to develop a sentiment analysis model to analyze the sentiment of tweets during international conflict. The trained model achieved high accuracies in predicting the sentiment of the tweets. The experiments indicate that the model effectively analyzes the sentiment of tweets related to the conflict.

\begin{figure}[!ht]
	\includegraphics[scale=.44]{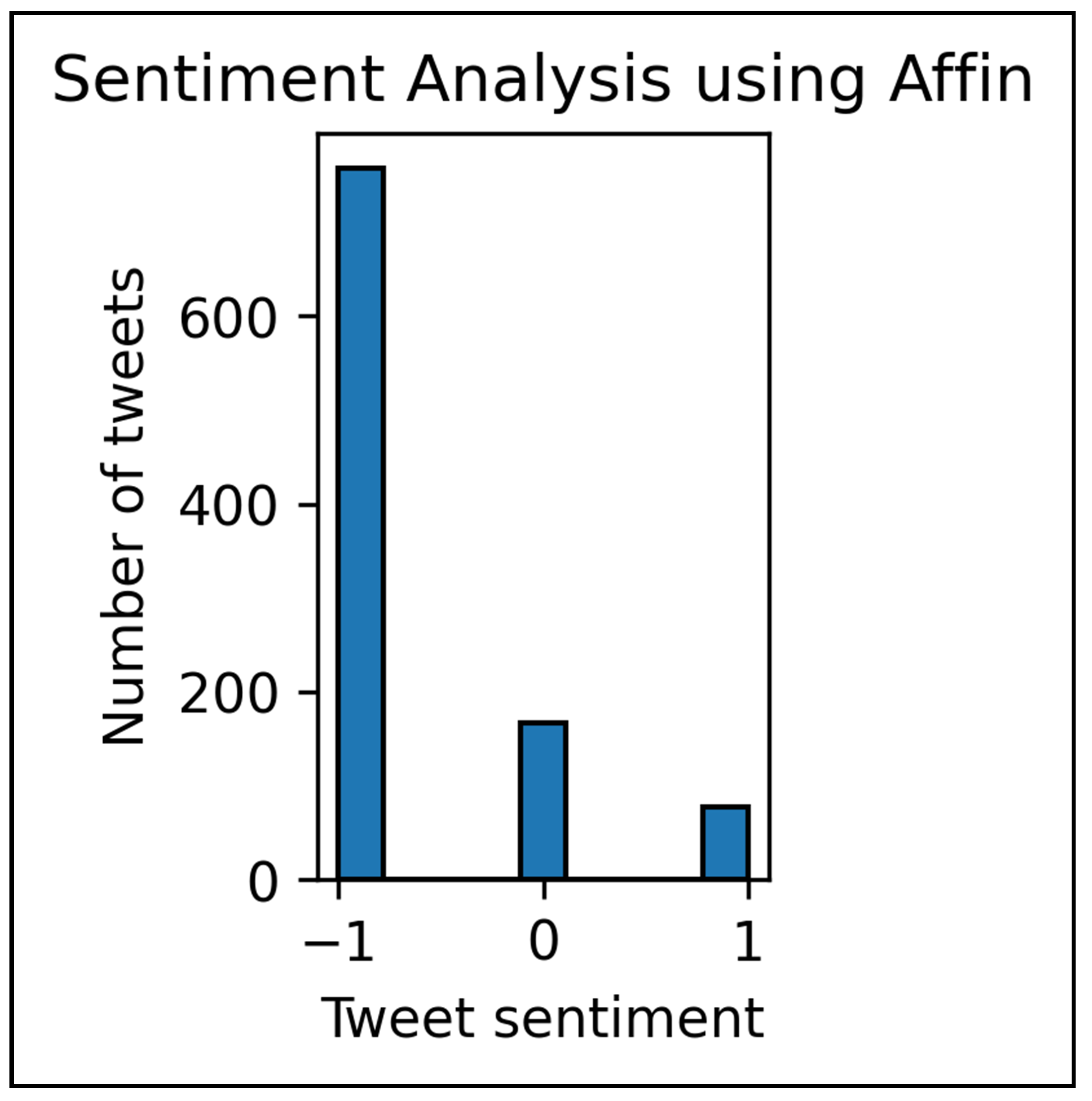}
	\centering
	\caption{Sentiment Analysis Results using Affin}
	\label{fig:affins}
\end{figure}

Figure~\ref{fig:affins} shows the number of tweets analyzed and the corresponding sentiment scores generated by the Affin lexicon. The sentiment scores range from -1 (most negative) to +1 (most positive). The blue bars represent the number of tweets.

\begin{figure}[!ht]
	\includegraphics[scale=.46]{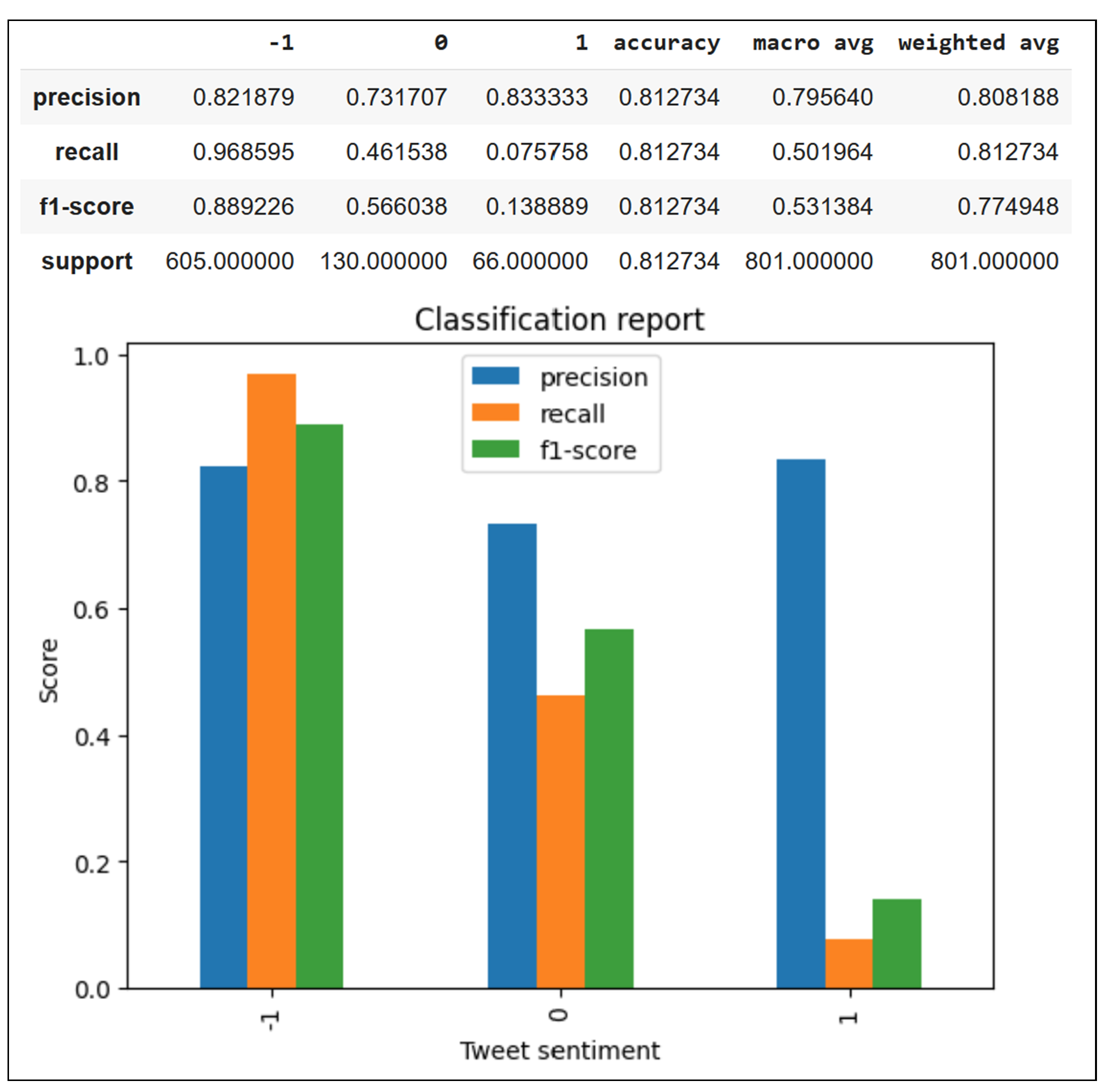}
	\centering
	\caption{Classification accuracy using Naïve Bayes}
	\label{fig:accnb}
\end{figure}

Figure~\ref{fig:accnb} represents the classification accuracy of around 80\% along with the metrics like precision, recall, f1-score. 

\begin{figure}[!ht]
	\includegraphics[scale=.42]{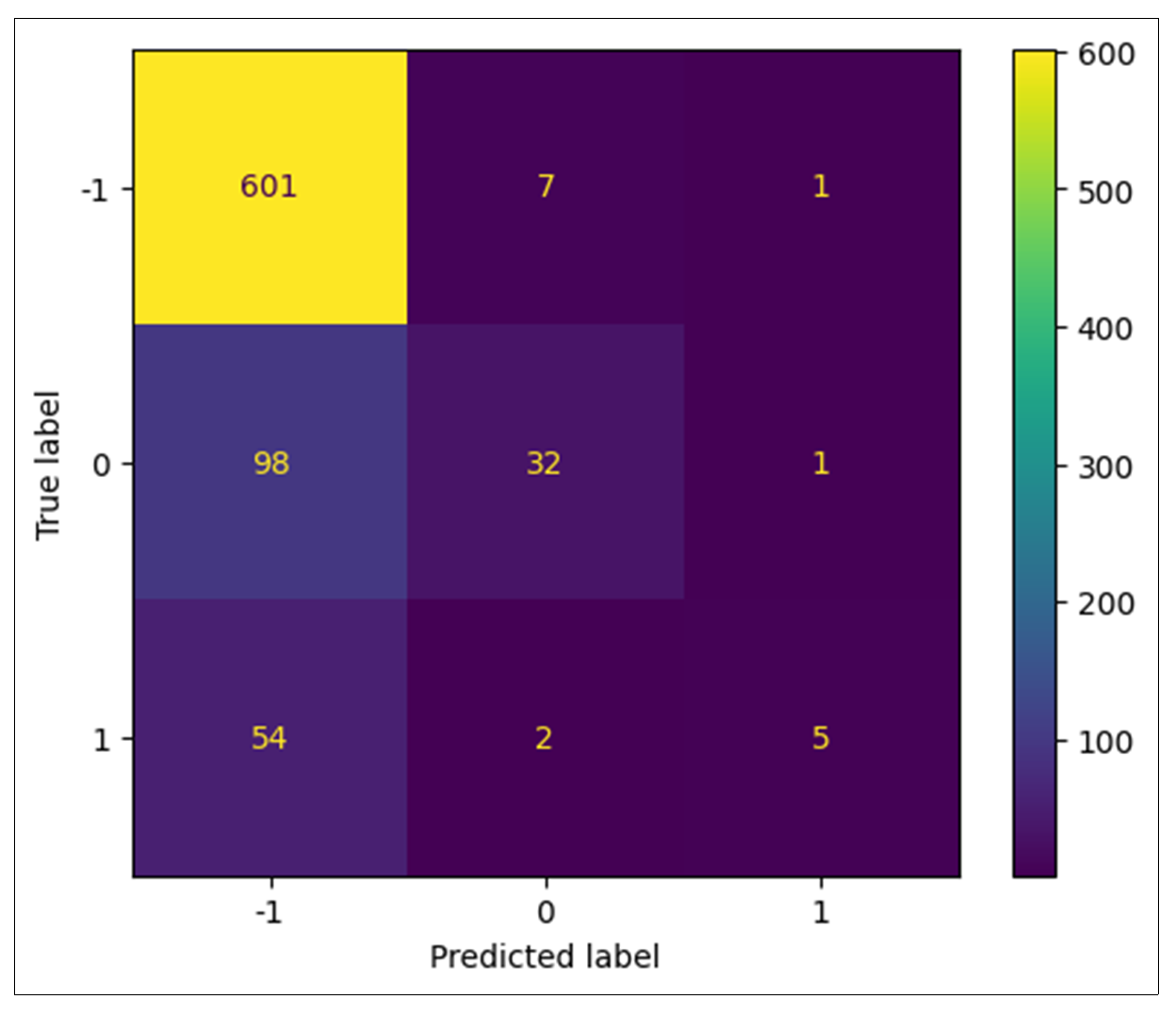}
	\centering
	\caption{Heatmap generated by the Multinomial Naive Bayes classifier}
	\label{fig:htp}
\end{figure}

Figure~\ref{fig:htp} shows the heatmap for the NB classifier, presenting the true positive (TP), true negative (TN), false positive (FP), and false negative (FN) classifications of the sentiment analysis model on the test dataset. The X-label is the Predicted value and the Y-label is the true or actual value. The Diagonal one (601,32,5) are the True Positive Values.

\begin{figure}[h]
	\includegraphics[scale=.34]{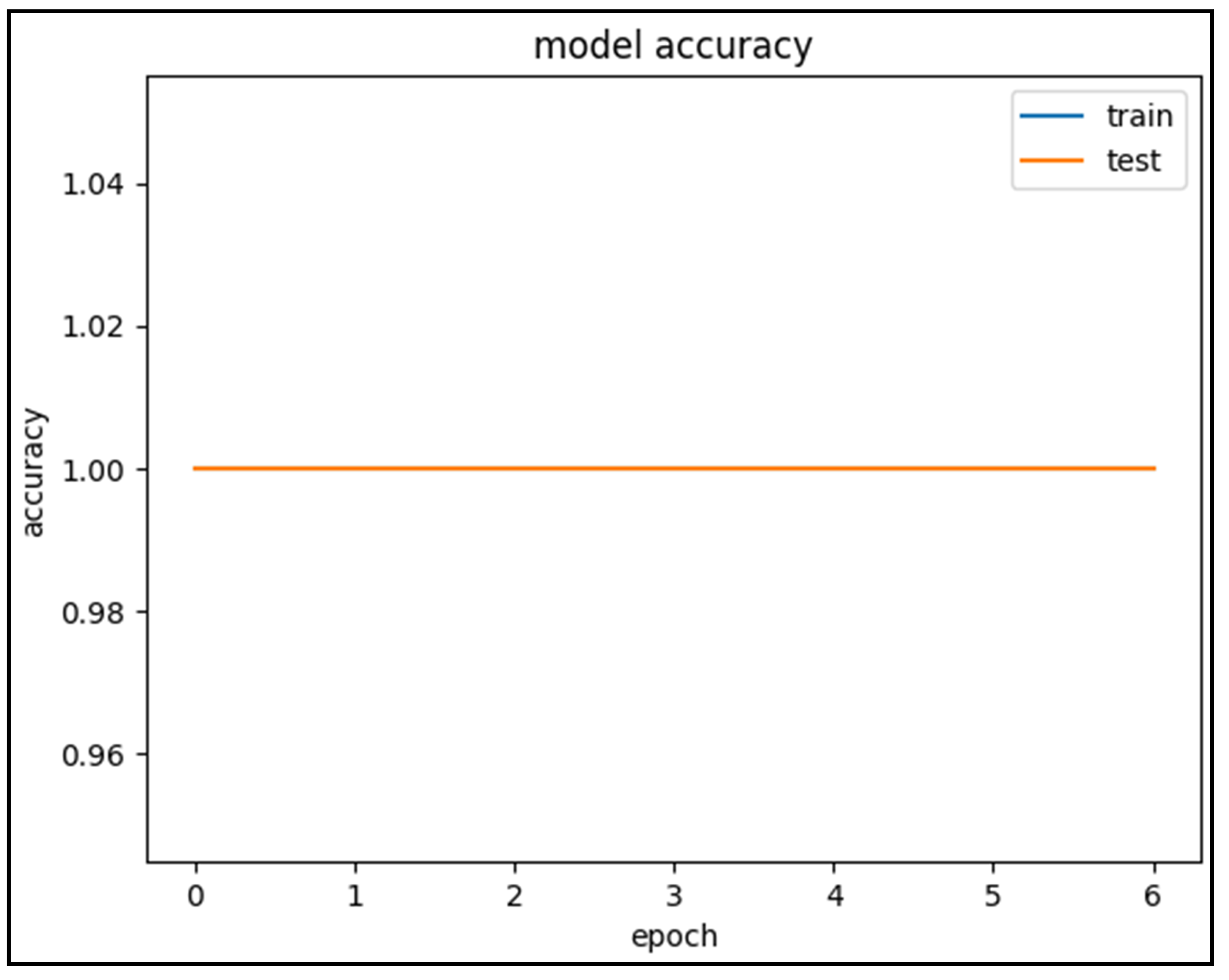}
	\centering
	\caption{Classification accuracy using NN}
	\label{fig:accnn}
\end{figure}

Figure~\ref{fig:accnn} represents the perfect classification accuracy by the NN model. While experimenting with the NN model, 100\% accuracy was achieved in classifying the Twitter data.

\begin{figure}[!ht]
	\includegraphics[scale=.36]{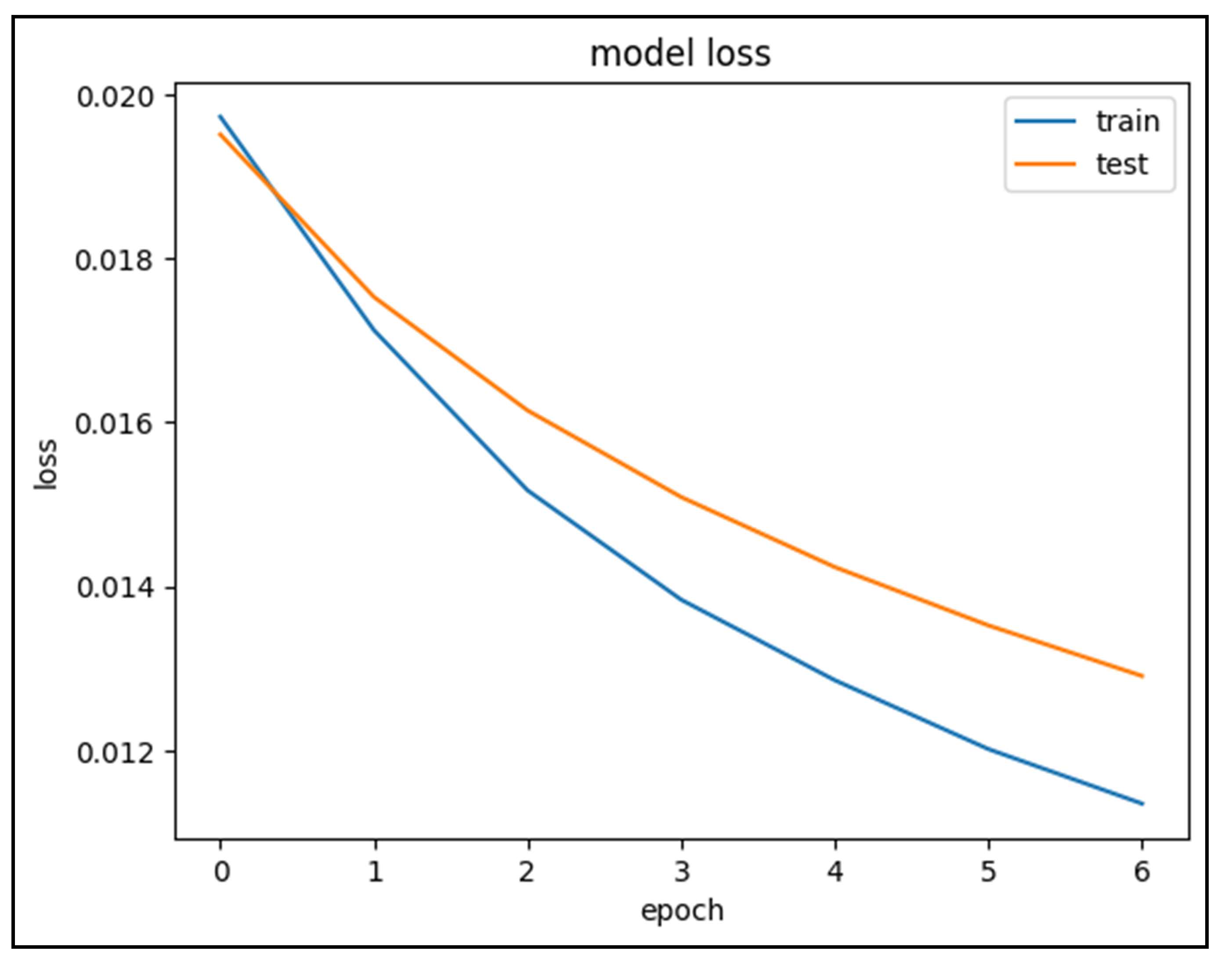}
	\centering
	\caption{Training and testing model loss of the neural network model}
	\label{fig:losnn}
\end{figure}

Figure~\ref{fig:losnn} 7 presents the training and testing model loss of a neural network model over multiple epochs. This can help you monitor the model’s performance and identify potential issues such as overfitting or underfitting. Various performance metrics such as precision, recall, and F1 score are presented to better understand the model’s performance. 

Our model achieved a precision of 90\%, recall of 80\%, and an F1 score of 85\% for the positive class and a precision of 83\%, recall of 92\%, and an F1 score of 87\% for the negative class. These metrics indicate that the model performs well for both positive and negative classes.

The experiments show promising results in finding and classifying sentiments about the war tweets.

\section{Conclusion}\label{con}
Our study contributes to the development of sentiment analysis models for analyzing social media data related to various events and topics. After thorough experiments with NB and NNs classifiers with the scrapped Twitter data, we identified that majority of the peoples' sentiment about any war time scenarios are negative. The second contribution of our study is to develop a twitter dataset for the researcher community on global conflicts. The findings can be useful for researchers interested in understanding public sentiment about the war and conflict expressed on the social media. The results of this research could have significant implications for understanding the impact of social media about public opinion and sentiment during conflicts. Sentiment analysis of social media data can provide valuable insights about public opinion, which can be used in making policies, decisions and communication strategies. Overall, this study demonstrates the effectiveness of using sentiment analysis models to analyze social media data related to major events and crisises. It opens up avenues for further research into the impact of social media on public sentiment during times of conflict and the development of more sophisticated sentiment analysis models. A larger dataset can be analyzed in future studies for more accurate results.

\balance
\bibliographystyle{IEEEtran}
\bibliography{CSCI2023Twitter-Camera-Ready.bib}

\end{document}